\RequirePackage{fix-cm}
\documentclass[smallextended]{svjour3}       
\smartqed  
\usepackage{graphicx}

\usepackage{tabularx}
\usepackage{caption}
\usepackage{graphicx}
\usepackage{amsmath}
\usepackage{amssymb}
\usepackage{hhline}
\usepackage[utf8]{inputenc}
\usepackage[dvipsnames]{xcolor}
\usepackage[breaklinks=true,colorlinks=true,
linkcolor=ForestGreen	,urlcolor=Blue,citecolor=OliveGreen,
bookmarks=true,bookmarksopenlevel=2]{hyperref}
\usepackage{rotating}
\usepackage{subfigure}
\usepackage{multirow}
\usepackage[ruled,vlined]{algorithm2e}
\usepackage{url}
\usepackage[justification=justified,singlelinecheck=false]{caption}
\usepackage{colortbl}
\usepackage{enumitem}
\usepackage{authblk}
\usepackage{placeins}
\usepackage{float}

\usepackage{geometry} 

\begin{document}

\newcommand{\newc}{\newcommand}
\newc{\be}{\begin{equation}}
\newc{\ee}{\end{equation}}
\newc{\ba}{\begin{eqnarray}}
\newc{\ea}{\end{eqnarray}}
\newc{\bea}{\begin{eqnarray*}}
\newc{\eea}{\end{eqnarray*}}
\newc{\ie}{{\it i.e.\ }}
\newc{\eg}{{\it e.g.\ }}
\newc{\etc}{{\it etc.\ }}
\newc{\etal}{{\it et al.}}

\title{SalSum: Saliency-based Video Summarization using Generative Adversarial Networks}


\titlerunning{SalSum: Saliency-based Video Summarization using Generative Adversarial Networks}        

\author[1]{George Pantazis$^\dagger$}
\author[1]{George Dimas$^\ast$}
\author[1]{Dimitris K. Iakovidis$^\star$}
\affil[1]{Department of Computer Science and Biomedical Informatics, University of Thessaly, Lamia, Greece 
}
\authorrunning{G. Pantazis et al.} 

\institute{$^\dagger$gpantazis@uth.gr, $^\ast$gdimas@uth.gr, $^\star$diakovidis@uth.gr}

\date{Received: Novermber 20, 2020 / Accepted:  \\
Multimedia Tools and Applications}

\maketitle
\begin{abstract}
The huge amount of video data produced daily by camera-based systems, such as surveilance, medical and telecommunication systems, emerges the need for effective video summarization (VS) methods. These methods should be capable of creating an overview of the video content. In this paper, we propose a novel VS method based on a Generative Adversarial Network (GAN) model pre-trained with human eye fixations. The main contribution of the proposed method is that it can provide perceptually compatible video summaries by combining both perceived color and spatiotemporal visual attention cues in a unsupervised scheme. Several fusion approaches are considered for robustness under uncertainty, and personalization. The proposed method is evaluated in comparison to state-of-the-art VS approaches on the benchmark dataset VSUMM. The experimental results conclude that SalSum outperforms the state-of-the-art approaches by providing the highest f-measure score on the VSUMM benchmark.
\end{abstract}


\section{Introduction} 

The structure of a video can be considered as a set of non-overlapping {\it scenes}, which are composed of semantically correlated {\it shots}. Each shot is composed of a sequence of video {\it frames} representing a continuous action. For example, a scene describing an office, may consist of shots describing actions of a clerk, such as typing and walking. In some cases a scene and a shot may be identical if the length of the first is small.

Videos are often lengthy in duration with complex content, including duplicate video contents or unuseful information regarding specific application contexts. Additionally, considering the huge amount of data that is transferred on a daily basis throughout various online platforms, the need of summaries describing their content becomes crucial. In many cases removing redundant or unuseful content can be beneficial, \eg to shorten its duration or focus only on important video segments. To this end, various video summarization (VS) approaches have been proposed throughout different application contexts such as in medicine, where the time of diagnosis can be drastically reduced \cite{iakovidis2010reduction}; in entertainment, where video selection can be based on a thumbnail or a movie trailer respectively \cite{DBLP:journals/mta/ShengCLL18}; in surveillance for security, in order to avoid hours of footage \cite{SenthilMurugan2018} \etc

VS methods can be categorized, based on their output, as static, where we obtain {\it keyframes} \cite{de2011vsumm}, or dynamic where summaries are described as {\it video skims} \cite{gygli2014creating}. In detail, a keyframe is a video frame which can be regarded as the most useful/representative frames of a video. On the other hand, as video skims, can be described as short video segments usually obtained by montaging frames before and after keyframes. The number of keyframes or the duration of the video skims is usually selected by using a threshold value. Another categorization can be considered regarding their execution, \ie whether is offline \cite{de2011vsumm}, \cite{gygli2014creating}, or online \cite{kim2019video}, with the majority of them belonging to the former. VS methods can also be categorized as supervised \cite{panda2017weakly} or unsupervised \cite{de2011vsumm}, \cite{gygli2014creating}. Supervised VS methods require a training set for learning patterns from previously annotated videos, whereas the unsupervised methods do not require any training, which generally makes them context independent.

Most of the VS methods incorporate feature extraction, which is a process for information abstraction enabling discriminative content representation. Such representations may describe spatial or temporal properties of a video, or both. Commonly used spatial features include video frame intensity, color, shape, and texture whereas temporal features include motion and orientation of consecutive frames. A recent trend to feature extraction suggests the use of machine learning, \eg via convolutional neural networks (CNNs) enabling automatic feature extraction. However, even though such feature extraction approaches can be very effective, they are usually computationally demanding and their physical interpretation is not straightforward. Such methods are usually combined with unsupervised clustering methods including among others k-means \cite{de2011vsumm}, \cite{lee2017robust}, \cite{kumar2018eratosthenes}, \cite{lu2017unsupervised}, fuzzy c-means (FCM) \cite{asadi2012video}, k-medoids \cite{singh2019pics}, \cite{DBLP:journals/mta/LuxMSBL10}, non-negative matrix factorization (NMF) \cite{iakovidis2010reduction}, \cite{DBLP:journals/mta/LiuSZF19}.

An additional beneficial approach in summarizing videos is that of visual attention as a result of visual saliency. Visual saliency indicates locations or proportions of pixels within images or frames that attract human attention. In other words, visual saliency is how humans would observe these specific images or frames without any filter in a short period of time. It is often categorized in bottom-up and top-down visual saliency \cite{ishtiaq2017visual}. The bottom-up visual saliency is based on the human stimulus, \ie a stimulus-impelled signal which is affected by features such as color, contrast, intensity, orientation \etc while the top-bottom visual saliency is a user-impelled signal, \ie depended on the user and affected by factors such as the appearance of a scene in a frame.

In order to assess the performance of VS methods, and compare them among other VS methodologies, we usually test them on specific benchmark datasets available in literature. Among these datasets include the OV dataset \cite{OVP} provided by the Open Video Project, the VSUMM dataset \cite{de2011vsumm}, \cite{VSUMM} consisting on videos collected through the OVP, the SumMe dataset \cite{gygli2014creating}, \cite{SumMe}, the Youtube dataset provided by \cite{de2011vsumm}, the TVSum dataset \cite{DBLP:conf/cvpr/SongVSJ15}, \cite{TVSum}. In the scope of this work, we focus on the VSUMM dataset since it consists of videos of different genres and it’s been broadly adopted for evaluation purposes by various methods in the literature.

In this paper we propose SalSum, a novel VS method based on the human eye fixation saliency information, estimated by a data-driven method. In detail, this method is being trained by exploiting a GAN model to estimate accurate saliency maps \cite{DBLP:journals/corr/PanCMOTSN17}. To the best of our knowledge this is the first VS analysis using a trainable saliency generation approach. By using this model, we generate saliency maps for each video frame. Then, hue histograms of the video frames are calculated to estimate a static score, while optical flow calculation between saliency maps, generated from consecutive video frames, provide a spatiotemporal score. The static and spatiotemporal score, are fused together forming a final score. The final keyframes are obtained via a user-based technique in accordance with the computation of score distance between frames and by using the histogram intersection dissimilarity measure. SalSum can be considered as a unsupervised method, since the only training that is required regards the GAN model, which is trained on estimating visual saliency.
\\ \\
The main contributions of the SalSum method are:

\begin{itemize}
    \item We propose a novel unsupervised VS approach that exploits the capabilities of a data-driven saliency estimation model, trained to estimate saliency maps based on human eye fixations. A saliency property that is suitable for VS applications is that saliency in general, is preserved with respect to objects appearing in consecutive video frames. 
    \\
    \item We consider spatiotemporal features; a) the perceptual representation of colors within each video frame and b) the optical flow of the saliency maps. Any radical change in the optical flow is more likely to be associated with a change in the video content. Therefore, we obtain more relevant video summaries with respect to human perception. To the best of our knowledge, this is the first work using the optical flow via saliency for obtaining video summaries.
\end{itemize}

The rest of this paper is organized as follows; In Section~\ref{sec:relatedWork} we discuss previous related works on VS via saliency as well as VS approaches used to be compare with in our evaluation experiments. In Section~\ref{sec:methodology} we discuss the data-driven saliency estimation model that generates saliency maps used for video summarization and we analyze our methodology step by step. The results from the evaluated experiments are presented in Section~\ref{sec:results}. Finally, we conclude the main aspects of this paper and discuss future prospects concerning this work.

\section{Related Work}
\label{sec:relatedWork}
\subsection{Saliency-based Video Summarization}

There are various saliency-based methods for video summarization proposed over the last years. Most of them focus on a visual attention approaches computed by either spatial or temporal features while others consider hybrid features as well.

In \cite{DBLP:journals/jiis/JacobPLP17} a variety of image features are considered for saliency-based VS. These include color, shape, texture and constrast as spatial features, motion computed via optical flow and facial recognition to obtain a spatial, temporal and sematic saliency score respectively. These scores are fused into a global visual attention model where VS is achieved by selecting consecutive score peaks. In \cite{DBLP:journals/mta/DarabiG17} a personalized saliency-based VS method was proposed. In that method, a score was assigned to each video frame based on their audio, visual and textual characteristics. Spatiotemporal saliency scores were estimated for keyframe selection based on the standard Itti saliency model \cite{DBLP:journals/pami/IttiKN98} and fused with the aforementioned scores. The video frames were classified into semantic categories based on Scale-Invariant Feature Transform (SIFT) visual descriptors. Similarly, in \cite{ishtiaq2017visual} a fused saliency model was presented, including both spatial and temporal saliency computed by color, contrast, luminance and texture, shape, size and motion features. Another study \cite{DBLP:journals/tip/ZhangHJYC17} investigated event saliency VS where a spatiotemporal saliency model was introduced, considering motion, vision, concept and speech. These features were extracted by VGG CNN and then they were grouped into clusters using k-means. The results presented were promising; however, it was noted that it lacks efficiency when it comes to noisy videos or complex backgrounds and blurred foregrounds. A visual attention model for VS was introduced in \cite{ubeduddinnovel}. It was based on spatial, temporal and semantic saliency using intensity, color, contrast, orientation, motion, edges and facial recognition respectively. This approach has been proved suitable for summarization of sports videos.

In \cite{DBLP:journals/isci/MademlisTP18} a global saliency model was presented where VS was achieved using a salient dictionary where keyframes reconstruct the whole video and simultaneously represent its most salient parts. For this approach, were used both low features such as luminance, color (hue), optical flow, edges and mid features like SIFT and IDT (improved dense trajectories). By clustering these features using the Jenk’s Natural Breaks Optimization algorithm the final summaries were obtained.  However, dictionary-of-representatives approaches do not guarantee outlier inclusion.  Similarly, an extended work to \cite{DBLP:journals/isci/MademlisTP18} considered an alternative SVD-based frame saliency method and presented to \cite{DBLP:conf/icassp/MademlisTP18}. Specifically, the video frame saliency estimation term was replaced with one based on Regularized SVD-based Low Rank Approximation, taking advantage of the well-established correlation between midrange matrix singular values and salient regions. This method correspondingly, is preferred on activity videos. In \cite{DBLP:journals/mta/HuL18} was proposed a novel VS method by fusing the global and local importance based on multiple features and image quality. These features included visual attention (static visual saliency and temporal saliency), HSV features (hue, saturation, light), rule of thirds, contrast and directionality and image quality. By fusing the importance an optimization of the clips achieved an optimal VS. However, if the frames quality is low, the number of the aforementioned features was redefined. A spatiotemporal saliency-based model was presented in \cite{kwan2018efficient} in order  to detect anomalies and summarize low quality videos. The used features included color, motion and events in order to compute saliency scores, apply smoothing and group them into clusters via a temporal clustering based on color. This method was used in videos that include people and/or vehicle motion, traffic accidents, abnormal motion, leaving suspicious objects \etc An automatic VS approach suitable for smart glasses was presented in \cite{DBLP:journals/mta/ChiuLW18} where features such color, texture, motion and audio (speeches) as well as visual saliency are considered. Through a k-means clustering on RGB color features the final summaries are selected by an adjusted rand index (ARI). This method is applied mainly on egocentric videos both indoors and outdoors. A visual attention unsupervised VS method was presented in \cite{DBLP:journals/mta/SunZZW18} where  spatial and temporal features are extracted using the CNN and optical flow respectively. In addition, the saliency maps are derived and the final summaries are obtained using the Earth Mover’s Distance and histogram intersection similarity between them and the ground truth. In \cite{DBLP:journals/ejivp/EjazBMCM18} the considered features included color, multi-scale contrast extracted by a Gaussian pyramid instead of usual color contrast to create static visual attention, motion obtained through optical flow to create a dynamic visual attention and relative motion orientation. Afterwards, an efficient fusion based on these three attention values takes place and keyframes are extracted by using the Euclidean similarity. 

A spatial and motion saliency prediction model (SMSP) was introduced in \cite{DBLP:journals/tcsv/PaulS19} based on  eye tracker data since human eyes are able to track moving objects accurately. The amount of salient regions and object motions are the two important features to measure the viewer’s attention level for determining the keyframes. This was achieved by extracting features such as intensity (grayscale) and motion. However, the eye tracking data concerning each video was obtained by humans watching the video.

Several state-of-the-art methods for saliency map generation based on human eye-fixation exploit the capabilities of machine learning algorithms, such as CNNs. Novel architectures using ensemble of deep networks (eDN) \cite{DBLP:conf/cvpr/VigDC14}, dual CNN architectures \cite{DBLP:conf/cvpr/PanSNMO16}, deep spatial contextual long-term recurrent convolutional network (DSCLRCN) \cite{DBLP:journals/tip/LiuH18} have been deployed for the task of visual saliency generation. In DeepGaze \cite{kummerer2014deep}, DeepGaze II \cite{kummerer2015information} and SUSiNet \cite{Koutras_2019_CVPR_Workshops} selected features from CNN models are used for the construction of saliency maps based on human eye-fixation while SalGAN (Saliency Generative Adversarial Network) exploits the GAN properties for accurate saliency generation \cite{DBLP:journals/corr/PanCMOTSN17}.Visual saliency based on human eye-fixation, has been proven to enable the robustness of methodologies beyond the context of video summarization \cite{dimas2019obstacle}.

\subsection{Video Summarization Methods using the OV dataset}

When it comes to the OV dataset there are several methods over the years that their evaluation is based on this benchmark dataset. Specifically, in \cite{DBLP:journals/jodl/MundurRY06} a VS method was proposed via the Delaunay Triangulation clustering (DT) where each frame was represented by an HSV color histogram, each histogram was represented as a row vector and the vectors for each frame were concatenated into a matrix while principal component analysis (PCA) was used for dimension reduction. After that the Delaunay diagram was built and clusters were created based on the edges of this diagram. Finally, frames nearest to the cluster center were selected as keyframes. 

In STIMO (STIll and Moving video storyboard) \cite{DBLP:journals/mta/FuriniGMP10} an on-the-fly VS approach was introduced by clustering the matrix produced by each frame’s histogram in HSV. It was also used triangular inequality to avoid redudant distance computations. Finally, the pairwise distance of consecutive frames was computed and if it were greater than a threshold value, the number of clusters were obtained where keyframes were extracted from each cluster. In VSUMM \cite{de2011vsumm} a VS method was proposed by extracting color features (in HSV color space with 16 bins in each channel) and frames were grouped using a k-means clustering. The Euclidean distance of consecutive frames was measured and keyframes were selected given a threshold value on this specific distance. There are two variants of this method, \ie VSUMM1 and VSUMM2 where in the first one the keyframes are the centers of the clusters while in the latter the obtained keyframes are processed by merging similar frames concerning the distances between the initial keyframes. In SMFR (Sparse Modeling for Finding Representative objects) \cite{Elhamifar_2019_ICCV} is achieved a VS by formulating this problem as a sparse multiple measurement vector problem. In SD (VS via Sparse Dictionary) \cite{DBLP:journals/tmm/CongYL12} VS was formulated as a novel dictionary selection problem using sparsity consistency, where a keyframes dictionary was taken in such a way where the original video could be reconstructed by this representative dictionary. 

In VISON (Vıdeo Summarization for Online applications) \cite{DBLP:journals/prl/AlmeidaLT12} VS was approached on the compressed domain while the quantity and quality of the final summaries were controlled by users. In SFKD (Spatio-temporal Feature-based Keyframe Detection) \cite{DBLP:journals/prl/MartinB13} every frame was described as a set of local features. A framework was proposed in which involved local features comparison and obtained by consecutive frames while building a graph-based appoach on the locality of these features. After that, attain trial graph partitions were attained through spectral clustering and the final keyframes were associated with these partitions. In KBKS (Keypoint-Based Keyframe Selection) \cite{DBLP:journals/tcsv/GuanWLDF13} was proposed a shot VS in which frames were described by their local features through a keypoint-based framework. 

VSCAN \cite{DBLP:conf/iciap/MahmoudIG13} was proposed as a VS method where color and texture features were employed for each frame while a DBSCAN clustering was used in order to select final keyframes. The HDPS (High Density Peak Search) \cite{rodriguez2014clustering} was a VS method via clustering in which cluster centers decribed by a higher density than their neighbors and therefere the number of clusters is arbitrary. In OffMSR (Offline VS via Minimum Sparse Reconstruction) \cite{DBLP:journals/pr/MeiGWWHF15} the SD method was extended by reconstructing the original video sequence with the smallest possible number of keyframes. Both the cases for online and offline VS were examined while two variants of this method been proposed, \ie OffMSR¬a and OffMSRm where in the first the most average frame was obtained as the first keyframe while in the latter the first keyframe was the one with the maximum magnitude. In VSQUAL (VS by image QUALity assessment) \cite{DBLP:conf/ciarp/CirneP14} VS is achieved by extracting color features and using a Feature Similarity Index Matrix (FSIM) which was computed by both the phase congruency and the gradient magnitude where the first one measures the significance of local features while the second one keeps the information concerning the contrast of an image. Finally shots were segmented by finding local minima points between consecutive frames in a sequence of FSIM and keyframes were selected based on the highest value of similarity. 

The VRHDPS (Video Representation by High Density Peak Searching) approach \cite{Wu2017} is an extention of the HDPS method and a VS method was presented via clustering where the Euclidean distance was calculated between two candidate frames, their local density was computed and the minimum distance for each candidate frame was calculated. In \cite{DBLP:conf/cvpr/MahasseniLT17} was presented SUM-GAN, a VS method using adversarial Long short-term memory (LSTM). Both a supervised and an unsupervised approach of this method were studied, \ie SUM-GANsup and SUM-GANdpp respectively. In SCVS (Sparse Coding VS) \cite{DBLP:journals/ijon/LiYLM17} was proposed a shot boundary detection method by using sparse coding to learn a dictionary from the given video. This was based on the idea that different shots cannot be reconstructed using the learned dictionary. Shot boundary detection was achieved by minimizing the reconstruction loss, restricting the sparsity of the reconstruction matrix while preserving video’s structure. Finally, frames with the minimum norm distance in each shot were selected as keyframes. 

In PFCVS (Parameter Free Clustering VS) \cite{DBLP:conf/cvip/MishraS16} was presented a VS method via an optimal k-means clustering using the SD Validity Index. This index is based on a average scattering of the clusters while is a useful tool at measuring the homogeneity and compactness of the clusters. In \cite{MusselCirne2018}, the VSQUAL was  extended to the VISCOM (Video Summarization using color co-occurrence matrices) method. The video frames were represented by the mean values of color co-occurance matrices, a shot boundary detection was achieved by the normalized sum of squared differences resulting in the final summaries of the video. In \cite{Kumar2018} was presented the ESVS (Eratosthenes Sieve based VS) method in which color features, spatial frequency and spectral residual were extracted while an optimal k-means clustering was used by selecting 5 videos. This method is based on the Eratosthenes Sieve idea and thus, sets of prime-number frames were used. In addition, there were two variants of this approach, \ie AVS and EVS selecting all frames and equal frames partition respectively. 

In \cite{martins2018opfsumm} OPFSumm (Optimum-Path Forest Summarization) was proposed where the optimum path forest clustering was used with k-neighborhood method on color, spectral and temporal information obtained by each frame. The MSKVS (Mean Shift-based Keyframe VS) method proposed in \cite{DBLP:journals/jvcir/HannaneEA18} aims to achieve VS by using the mean-shift clustering algorithm on visual and temporal features of the video frames obtained by the scale-invariant feature transform (SIFT), the difference of Gaussian (DoG), the orientation and the entropy. In PICS (Parameter Independent Clustering Strategy) \cite{singh2019pics} a parameter-free VS was presented via k-medoid clustering on the RGB color features of video frames. In addition, the Calinski–Harabasz index (CH index) was used enabling users to select keyframes without any computational cost. In \cite{DBLP:journals/jifs/MohanN19} SAEKMVS (Sparse AutoEncoders and K-Means clustering VS) was presented where VS was achieved by the Dual Layer Loopy Belief Propagation Network (DLBPN) and a k-means clustering. In KEGC (Keyframe Extraction via Graph Clustering) \cite{DBLP:journals/sivp/GharbiBZ19} was introduced a VS approach using graph clustering after local features were extracted from video frames using the LBP descriptor and points of interest using the SIFT algorithm. In CENTRIST VS \cite{DBLP:journals/mta/BendraouES19} a VS method was proposed via singular value decomposition (SVD) of centrist features (census transform histogram), which is a visual descriptor for recognizing places or scene understanding. Final keyframes were selected by a k-means clustering on the extracted features.

Although several VS approaches have been proposed, many of which have been based on visual attention and saliency models, to the best of our knowledge none of them has considered the optical flow of saliency maps estimated between consecutive frames. In this paper, unlike previous studies, we investigate this novel approach using a saliency estimator trained on human eye fixations. The rationale for this approach relies on the observation that saliency is generally preserved with respect to the objects presented in consecutive video frames; therefore, a change in the optical flow of the saliency is likely to be associated with a change in the video content. Also, by using a generative model based on human eye fixations along with a perceptual representation of colors, we expect to obtain more perceptually relevant video summaries.

\section{Proposed Methodology}
\label{sec:methodology}

\begin{figure*}[!h]
\centering
\includegraphics[width = 0.85\textwidth]{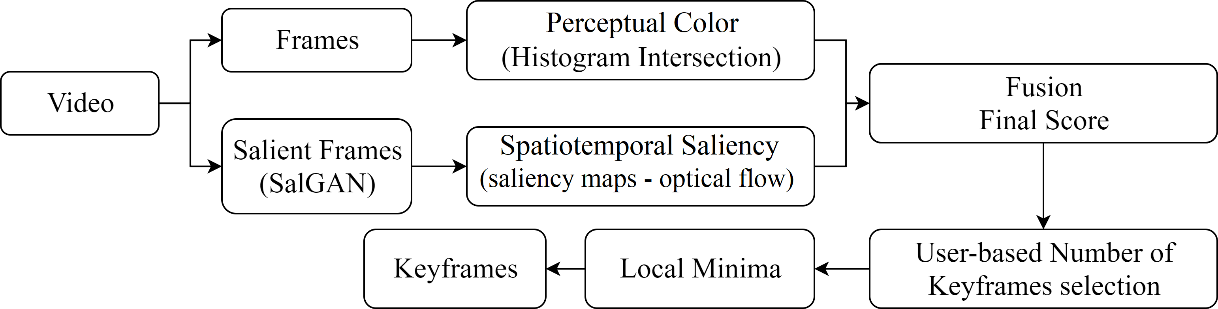}
\caption{The proposed SalSum methodology.}
\label{fig: fig1}
\end{figure*}

The saliency-based summarization methodology proposed in this work, named SalSum, is schematically illustrated in Fig.~\ref{fig: fig1}. The method considers that a user selects a prefered number of keyframes, constituting a video summary. Each video frame is analysed both in terms of color and saliency. Color is represented by hue histograms, and saliency maps are estimated by the application of SalGAN \cite{DBLP:journals/corr/PanCMOTSN17}. Then, a static score is estimated from the similarity of the hue histograms, and a temporal score is computed based on the optical flow between consecutive saliency maps estimated by SalGAN given the respective RGB input frames. These two saliency scores are fused together to obtain a final saliency score in which any radical change may indicate a scene transition. The keyframes are selected by considering the local minina of the signal formed by this score over time. The details of this method are provided in the following subsections. 

\subsection{Saliency Maps Estimation}

Several state-of-the-art methods exist for saliency map generation using CNN models. It is worth mentioning that any of the methods mentioned above provide us with reliable saliency maps. However, the comparison of these methods exceeds the main purpose of this study. Hence, for this work, SalGAN has been selected for the task of visual saliency map generation. SalGAN is an end-to-end saliency map generation model and among the best performing model in saliency estimation. SalGAN produces saliency maps with an accuracy comparable to that of other top-performing models in the saliency generation domain.

SalGAN (Saliency Generative Adversarial Network) is a state-of-the-art data-driven saliency prediction method \cite{DBLP:journals/corr/PanCMOTSN17}. Specifically, SalGAN is a generative adversarial network (GAN) trained with a joint loss function consisting of two losses, a recostructional and an adversarial loss.  The binary cross-entropy has been chosen as a reconstructional loss. The network is trained to produce saliency maps, predicting the human eye-fixations on images, \ie where the visual attention of humans is focused on images. Since SalGAN, is a GAN, two networks are required during its training, a generator and a discriminator network. The generator model is a deep convolutional neural network with an encoder-decoder architecture, while the discriminator is a simple convolutional network with a fully-connected part. The purpose of the generator is to generate accurate saliency maps, predicting the human eye-fixation on the input image. The goal of the discriminator is to distinguish the generated saliency maps from ground truth data. These two networks are trained in parallel, so the generator constantly trying to ``fool" the discriminator, while the discriminator is trying to get better into classifying correctly the ground truth and the generated images. This procedure can be described as a min-max game between the generator and the discriminator, where the minimization of the generator loss, maximizes the loss of the discriminator by producing accurate saliency maps, indistinguishable from the ground truth.

\begin{figure*}[!h]
\centering
\includegraphics[width = 0.9\textwidth]{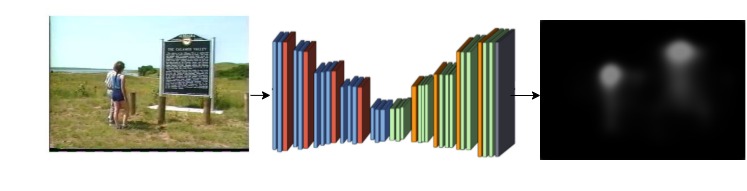}
\caption{The SalGAN architecture on frame 1200 of video \#25 of VSUMM. The  convolutional layers of the encoder are depicted withblue andthe pooling layers in red color. The convolutional layers of the decoder are in green andthe up-sampling layers in orange color while the the final layer used for activation is in gray}
\label{fig: fig2}
\end{figure*}

The encoder component of the SalGAN network has the same architecture to that of VGG-16 model \cite{simonyan2014very}, including both the convolutional and the max-pooling layers (colored in gray and purple in Fig.~\ref{fig: fig2} respectively). However, the final pooling layer and the fully connected (FC) part have been removed for the task of saliency map generation. The maxpooling layers enable the dimensionality reduction of the feature maps produced by the convolutional operations on the input data. The encoder was pretrained on ImageNet \cite{deng2009imagenet}. The encoder is tasked to produce a latent representation of the input image which is then used for the saliency map generation. The decoder part of SalGAN is used to generate a saliency map, given the output of the encoder. The architecture of the decoder is the same as the encoder, with the max-pooling layers replaced with up-sampling layers, and the convolutional layers in reverse order (Fig.~\ref{fig: fig2}). The activation function used in all convolutional layers of the is the Rectified Linear Unit (ReLU), with an exception to the last convolutional layer of the decoder where the sigmoidal function (logsig) is employed. Fig.~\ref{fig: fig2} illustrates the SalGAN architecture for the saliency map generation given an input image. The  convolutional layers of the encoder are illustrated in blue, the maxpooling layers in red while the decoder’s convolutional layers are in green and the up-sampling layers in orange. The final convolutional layer is depicted with gray color.

\subsection{Feature Extraction and Score Estimation}

The proposed methodology considers features extracted from both the spatial and temporal domain. Spatial features include perceptual color, visual saliency, and optical flow used to assess the temporal changes of saliency over time; therefore, in that sense, it encodes spatiotemporal saliency.

\subsubsection{Perceptual color}

Throughout this paper we select the color in the spatial domain by computing the quantized, $n$ bins , hue histograms, where $n = {4,8,16,32,64}$. TThis selection was based on the perceptual nature of the HSV color space, which resemblance how humans interpret color. In detail, in HSV space, color is dependent to the lighting conditions of a scene which reflects the how humans perceive it. Furthermore, the hue channel describes exclusively color and thus, it is suitable for color feature extraction. In addition, when a perceptually uniform color space is selected it may be appropriate to use uniform quantization \cite{niranjanan2012performance}. Hence, we compute the consecutive hue histograms, quantized in n bins, and then we measure their divergence using the normalized histogram intersection dissimilarity distance which is defined as follows \cite{niranjanan2012performance}:
\be 
d = 1 - \sum_{i=1}^N \frac{\text{min}\left(\text{hist}_i, \text{hist}_{i+1}\right)}{\text{max}\left(\text{hist}_i, \text{hist}_{i+1}\right)}
\label{eq:distance}
\ee
where $N$ is the total number of salient frames used in the analysis and hist$_i$, hist$_{i+1}$ are the $i-$th hue histogram and its consecutive. The smaller the value of $d$, the more similar the consecutive frames are.

\subsubsection{Spatiotemporal saliency}

In the spatiotemporal domain, we compute the normalized motion intensity between the saliency maps of the consecutive video frames using optical flow. This is estimated by the intersection divided by the union of the saliency maps, \ie their minimum over their maximum values. The optical flow is calculated between the saliency maps of consecutive frames of a video $V$, at $t$ and $t + \Delta t$ respectively. A pixel {\bf $p$} of a frame, at time $t$ of a video $V$, can be described as {\bf $p$}$= (x, y, t)$ and its intensity value as $V(\textbf{$p$})$. The corresponding pixel {\bf $p$}$^\prime = (x + \Delta x, y + \Delta y, t + \Delta t)$, on the next frame at time $t + \Delta t$ will have an intensity of $V(\textbf{$p$}^\prime)$. Assuming that the intensity will be conserved between the two frame, the following relation is derived:
\be V(\textbf{$p$}) =  V(\textbf{$p$}^\prime)
\label{eq:eq2} \ee
Considering that the movement between these frames is negligible, the Taylor expansion of Eq.~\eqref{eq:eq2} will give us
\be
\frac{\partial V}{\partial x}\frac{\Delta x}{\Delta t} + \frac{\partial V}{\partial y}\frac{\Delta y}{\Delta t} + \frac{\partial V}{\partial t} = 0 \ \Rightarrow \ V_x U_x + V_y U_y + V_t = 0 \ \Rightarrow \ \overrightarrow{\nabla} V \cdot \overrightarrow{U} = - V_t
\label{eq:eq3}
\ee
where $V_x = \partial V / \partial x, V_y = \partial V / \partial y$ and $U_x = \Delta x / \Delta t, U_y = \Delta y / \Delta t$ are the velocities along the $x-y$ plane.
\\ \\
However, Eq.~\eqref{eq:eq3} has two unknown variables; thus, it cannot be solved. There are several approaches available in the literature. In this paper, we consider the Lucas-Kanade optical flow method \cite{lucas1981iterative}. It should be noted that KLT uses the spatial intensity information to guide search towards the direction that secures optimal matching and is faster to several other aproaches since it checks for significantly fewer potential matches among images \cite{dimas2017intelligent}. This method assumes that the flow field is spatially preservative, \ie the pixels within a certain window are assumed to have identical value of velocity. Therefore, if we consider a point {\bf $p$} and a window centered at this point all the N pixels within this window must follow Eq.~\eqref{eq:eq3}. Specifically we have

\ba
& V_x(q_1) U_x + V_y (q_1) U_y &= - V_t(q_1) \nonumber \\
& V_x(q_2) U_x + V_y (q_2) U_y &= - V_t(q_2) \nonumber \\
  & \vdots & \vdots \nonumber \\
& V_x(q_N) U_x + V_y (q_N) U_y &= - V_t(q_N)
\ea

This problem is of the form $Av = B$, where
\be
A = 
\begin{pmatrix}
V_x(q_1) & V_y(q_1) \\
V_x(q_2) & V_y(q_2) \\
\vdots & \vdots \\
V_x(q_N) & V_y(q_N)
\end{pmatrix}, \ \ \ v =
\begin{pmatrix}
U_x \\ U_y
\end{pmatrix} \ \ \text{and} \ \ B = -
\begin{pmatrix}
V_t(q_1) \\ V_t(q_2) \\ \vdots \\ V_t(q_N)
\end{pmatrix}
\label{eq:eq5}
\ee

This gives an over-determined linear system which can be solved using a least squares method enabling us to compute the velocity vectors $U_x$, $U_y$ as $U = (A^T A)^{-1} A^T B$, where $A^T$ is the transpose matrix of $A$. 

\begin{figure*}[!h]
\centering
\includegraphics[width = 0.9\textwidth]{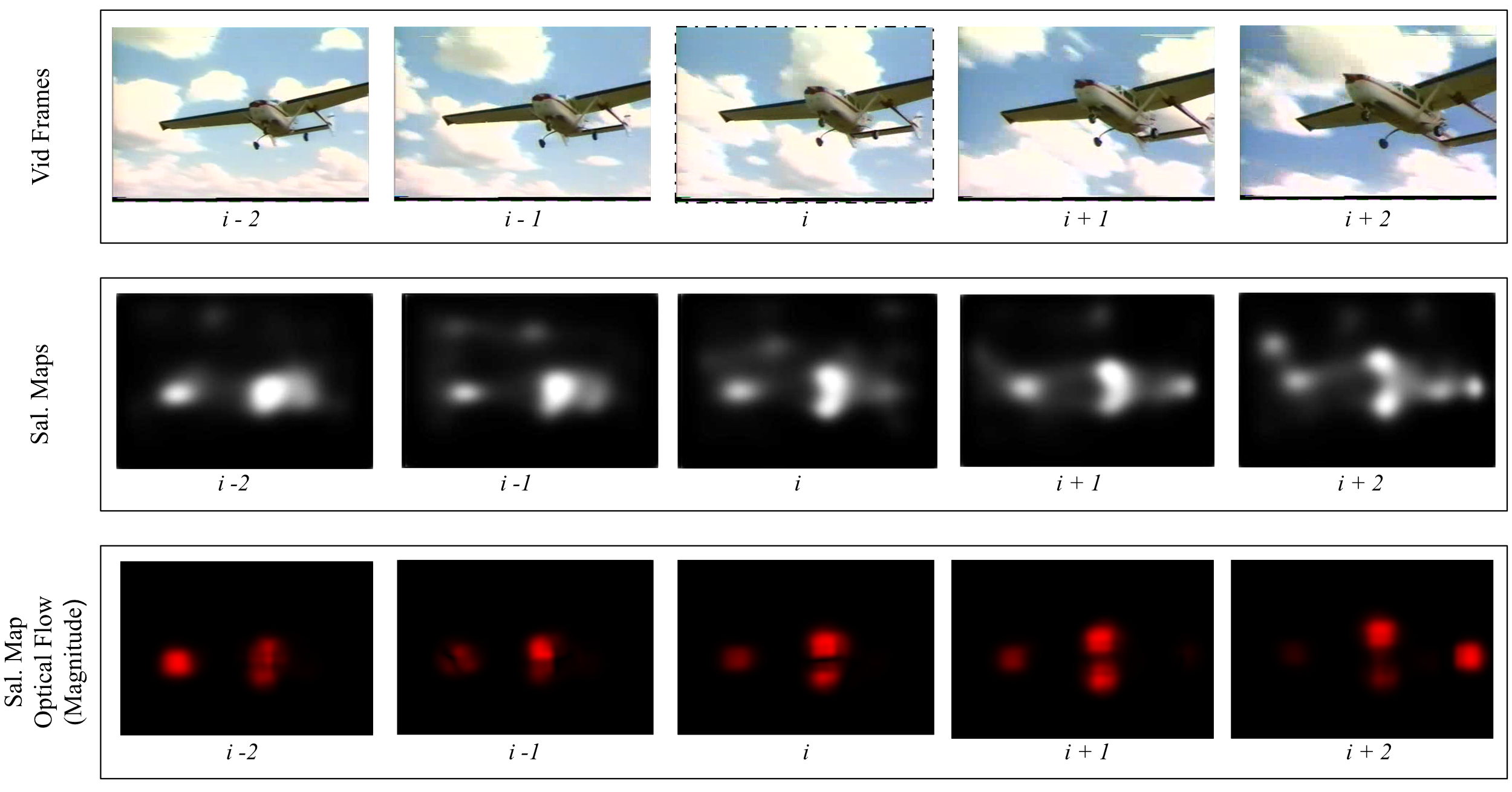}
\caption{A visual representation of the Optical Flow estimation based on the saliency information, i points to the keyframe index.}
\label{fig: fig3}
\end{figure*}

Then, the motion intensity (magnitude) is computed as $\sqrt{U_x^2 + U_y^2}$ and is used for the temporal saliency score in our analysis. An example of this score is shown in Fig.~\ref{fig: fig3}.

\newpage
\subsubsection{Score fusion for keyframe extraction}
\label{subsubsec: finalscore}

Both the static and the temporal saliency scores are fused into a final saliency score as
\be
\text{FinalScore} = \sum_i \frac{S_i}{\text{variance}(S_i)}, \ \ i=\{1,2\}
\label{eq:score}
\ee
where $S_i$ represents the different scores considered. In this paper $S_1$ and $S_2$ are the static and temporal scores respectively. To maximize the quality of this score we also apply a smoothing filter to Eq.~\eqref{eq:score}.

\begin{figure*}[!h]
\centering
\includegraphics[width = 0.9\textwidth]{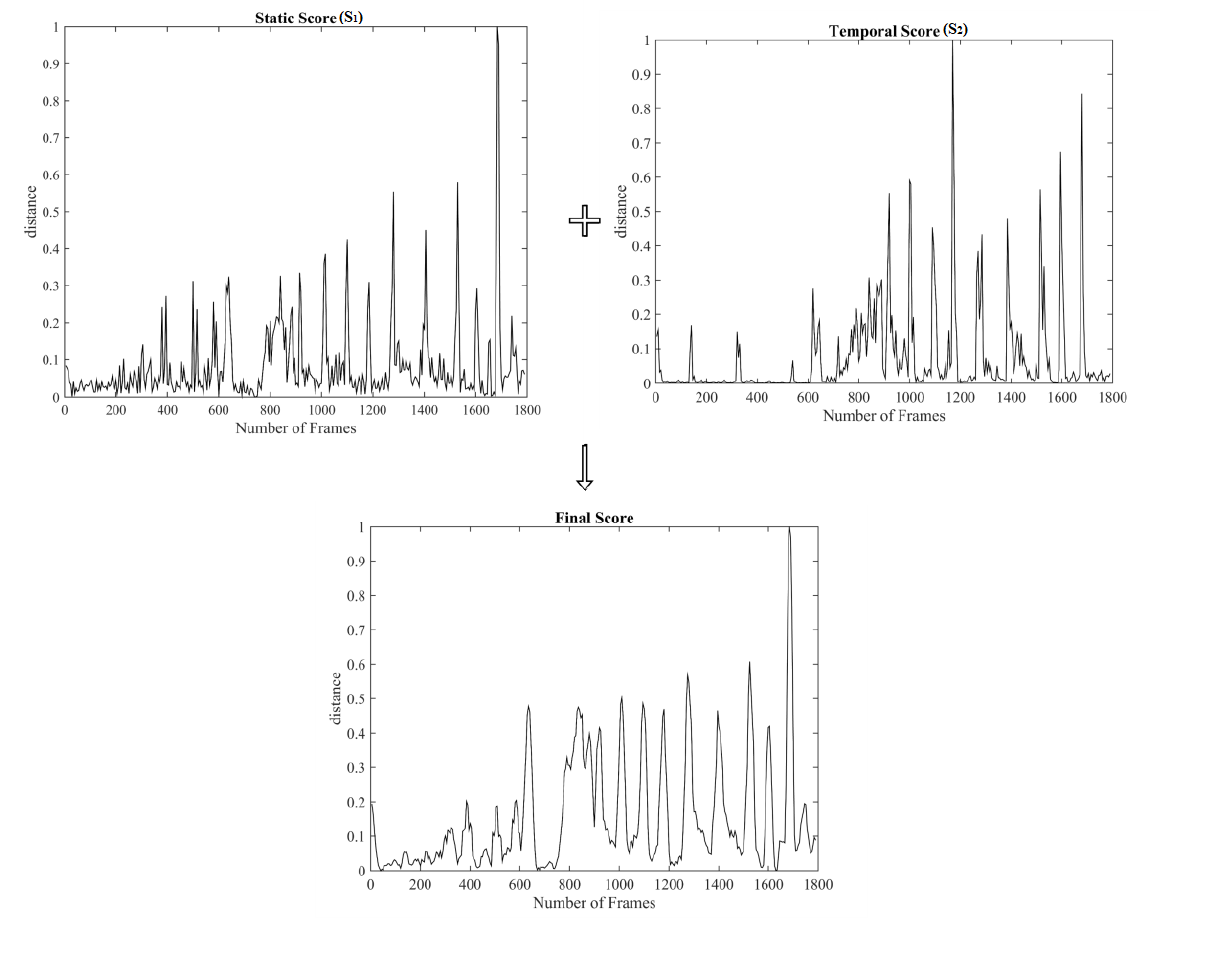}
\caption{Static, Temporal and Final Saliency scores for video \#25 of the VSUMM dataset.}
\label{fig: fig4}
\end{figure*}

Using Eq.~\eqref{eq:distance} we obtain the static saliency score $d$ of a video, \ie $S_1$. In addition, using Eq.~\eqref{eq:eq5} we compute the temporal saliency score, \ie $S_2$. In Fig.~\ref{fig: fig4}, an example of this process is illustrated, where we show both the static and the temporal saliency scores along with the final saliency score as a result of the fusion for the video \#25 of VSUMM dataset. Minima around peaks with non-negligible difference are candidates to be keyframes, as the probabillity to indicate different scenes is high.

\FloatBarrier

\section{Experiments – Results}
\label{sec:results}

\subsection{Experimental Setting}

The variaty of VS methods compared on specific datasets is quite large enabling us to test the effectiveness of SalSum. In this section we present our experimental setting and the performance evaluation process used in our analysis. Additionally, a comparison among the proposed SalSum and other VS methodologies is presented and discussed.

\subsubsection{Dataset}

We conduct our experiments on the public benchmark dataset VSUMM \cite{de2011vsumm}, \cite{VSUMM}. It is composed of 50 generic videos of several genres such as documentary, educational, ephemeral, historical, lectures \etc, which are selected from the open video project (OVP) \cite{OVP}. Each and every one of these videos are in MPEG-1 format with a 29.97 framerate value and a size of $352 \times 240$ pixels. Their duration lies between 1 to 4 min while the whole dataset consists of approximately 75 min of videos. Every video is watched and evaluated by 5 humans who selected ground truth (GT) frames according to their subjective opinions. Since this dataset is publicly available and is consisted of videos along with their respective summarization GT, its popularity is quite large and thus, several VS methods select it to perform their experiments \cite{Kumar2018}, \cite{singh2019pics}, \cite{DBLP:journals/jodl/MundurRY06}-\cite{DBLP:journals/mta/BendraouES19}.

\subsubsection{Evaluation}

The precision, recall and f-measure metric scores are employed in this work for the quantitative evaluation of the proposed methodology. The precision measure is defined as
\be
\text{precision} = \frac{N_\text{match}}{N_\text{candidate}}
\label{eq:prec}
\ee
and determines how many selected frames are relevant. $N_\text{match}$ is the number of the correct matches between the GT and automatic summaries obtained from any VS method and $N_\text{candidate}$ is the total number of the automatic summaries by a specific VS approach. This measures the optical comprehensiveness of a video, \ie the accuracy of the VS.

The recall can be described as a coherence metric of a video and determines how many relevant frames are selected. It is defined as
\be
\text{recall} = \frac{N_\text{match}}{N_\text{GT}}
\label{eq:recall}
\ee
where $N_\text{GT}$ is the total number of the GT frames provided by a user.

While these metrics describe well enough the summarization quality, the most common and used metric is the harmonic mean of both the precision and recall known as f-measure and defined as
\be
\text{f-measure} = \frac{2 \times \text{precision} \times \text{recall}}{\text{precision} + \text{recall}}
\label{eq:fmeasure}
\ee
In order to conclude which features contribute the most to our methodology, several approaches were compared. These include spatial features, \ie the perceptual hue color space with various histogram bin number, spatiotemporal features, \ie optical flow, and texture features individually and/or fused together by various fusion forms. These features concern either the video frames or the saliency maps generated by the corre-sponding frames or both. The intensity of the saliency maps was considered as well. The results of this comparison are summarized in (Table~\ref{tab:table1}).

This is a measure of the overall quality considering both precision and recall in its calculation. A high value of the f-measure signifies a higher overall accuracy of the evaluated methodology. Since we have 5 users per video and therefore 5 sets of GT, we use the mean f-measure in order to obtain one score per video.

\begin{table}[H]
\centering
\caption{Experiments on various features both individually and variance-based fused. These features concern either the video frames or the corresponding saliency maps.}
\label{tab:table1}
\begin{tabular}{lc}
\hhline{==}
\multicolumn{1}{c}{Feature}                                           & Score          \\ \hline
\multicolumn{1}{c}{\textit{Individual Features}}                      &                \\ \hline
Static (saliency maps)                                                & 0.728          \\
Optical flow (video frames)                                           & 0.776          \\
Optical flow (saliency maps)                                          & \textbf{0.806} \\
Hue8 (video frames)                                                   & \textbf{0.796} \\
Hue16 (video frames)                                                  & 0.728          \\
LBP (video frames)                                                    & 0.645          \\
LBP (saliency maps)                                                   & 0.675          \\ \hline
                                                        
\multicolumn{1}{c}{\textit{Variance-based fused Features}}            &                \\ \hline
Hue16 (video frames) + Optical flow (video frames)                      & 0.817          \\
Hue32 (video frames) + Optical flow (video frames)                      & 0.815          \\
Hue16 (video frames) + Static (saliency maps) + Optical flow (video frames) & 0.803          \\
Hue32 (video frames) + Static (saliency maps) + Optical flow (video frames) & 0.799          \\
Hue4 (video frames) + Optical flow (saliency maps)                         & 0.818          \\
Hue8 (video frames) + Optical flow (saliency maps)                         & \textbf{0.835} \\
Hue16 (video frames) + Optical flow (saliency maps)                        & 0.827          \\
Hue32 (video frames) + Optical flow (saliency maps)                        & 0.824          \\
Hue64 (video frames) + Optical flow (saliency maps)                        & 0.827          \\
Hue16 (video frames) + Static (saliency maps) + Optical flow (saliency maps)   & 0.771          \\
Hue32 (video frames) + Static (saliency maps) + Optical flow (saliency maps)   & 0.763          \\
HSV8 (video frames) + Optical flow (saliency maps)                         & 0.823          \\
HSV16 (video frames) + Optical flow (saliency maps)                        & 0.827          \\
HSV32 (video frames) + Optical flow (saliency maps)                        & 0.828          \\
HSV64 (video frames) + Optical flow (saliency maps)                        & 0.810          \\ \hhline{==}
\end{tabular}
\end{table}

\subsection{Results and Comparison with existing methods}

Initially, these features were examined individually in order to find the most efficient contribution for both the video frames and the corresponding saliency maps. The hue color space quantized to 8 bins gives better accuracy for video frames, whereas optical flow contributes the most to its saliency maps. After that, several feature combinations were applied using the variance-based fusion of Eq.~\eqref{eq:score}. Although, most combinations give similar results, the most efficient is the one of the individual features that give the most promising results, fused together.

The final fused score introduced in Section~\ref{subsubsec: finalscore} (Eq.~\eqref{eq:score}) was obtained after experimenting and comparing various fusion techniques for both the static and temporal features. However, we considered many fusion approaches such as the linear, arbitrary weights, min$/$max \etc \cite{evangelopoulos2013multimodal}, \cite{zlatintsi2012saliency} and we concluded that the variance-based fusion gives us better quality of the final summaries over the others (Table~\ref{tab:fusion}).

\FloatBarrier
\begin{table}[H]
\centering
\caption{Fusion techniques considered and their corresponding f-measures. The variance-based is preferable. We use as $S_1$ and $S_2$ the static and temporal scores respectively.}
\label{tab:fusion}
\resizebox{0.85\textwidth}{!}{%
\begin{tabular}{ccc}
\hline
Fusion         & \multicolumn{1}{c}{Form}                                                                                   & f-measure      \\ \hhline{===}
Linear         & $\sum_i w_i S_i$                                                                                           & 0.807          \\
Min            & min$(S_i)$                                                                                                 & 0.798          \\
Max            & max$(S_i)$                                                                                                 & 0.798          \\
Exponential    & $\sum_i (1-w_t)S_i, \ \ w_t = d e^{1-d}, \ \ d = \text{max}(\text{frame}_1) - \text{min}(\text{frame}_2))$ & 0.803          \\
Logarithmin    & $\sum_i \text{min}(S_i - w_i) + \text{max}(w_i), \ \ w_i =\log(1/\text{var}(S_i))$                         & 0.821          \\
Complex        & $\sqrt{S^2}, \ \ S = S_1 + i S_2$                                                                          & 0.826          \\
Harmonic Mean  & $2 \times \prod_i S_i / \sum_i S_i$                                                                        & 0.810          \\ \hhline{===}
Variance-based & $\sum_i S_i/\text{var}(S_i)$                                                                               & \textbf{0.835} \\ \hline
\end{tabular}%
}
\end{table}

In order to obtain more robust conclusions about the performance of the proposed method in comparison to the state-of-the-art methods, we have selected the VS methods that conduct their experiments on VSUMM. These methods were published in the timespan of 2006-2019 and were described in    Section~\ref{sec:relatedWork}. In Table~\ref{tab:comparison} we summarize the f-measure scores of these methods along with the one obtained by our methodology. The best score appears in boldface typesetting whereas the second best is underlined. It can be observed that the SalSum method provides an adequate summarization for the VSUMM dataset.

\begin{table}[H]
\centering
\caption{A comparison between the f-measure scores obtained by several VS methods using the VSUMM dataset and the one obtained by the SalSum method. Best score appears in boldface typesetting whereas second best is underlined.}
\label{tab:comparison}
\resizebox{0.55\textwidth}{!}{%
\begin{tabular}{cccc}
\hline
Method               & Year & f-measure         & Reference \\ \hhline{====}
DT                   & 2006 & 0.711             & \cite{DBLP:journals/jodl/MundurRY06}   \\
STIMO                & 2010 & 0.749             & \cite{DBLP:journals/mta/FuriniGMP10}   \\
OV                   & 2011 & 0.736             &   \cite{OVP}        \\
VSUMM1               & 2011 & {\underline{0.788}}          & \cite{de2011vsumm}          \\
VSUMM2               & 2011 & 0.772             &     \cite{de2011vsumm}      \\
SMFR                 & 2012 & 0.650             & \cite{DBLP:conf/cvpr/ElhamifarSV12}, \cite{DBLP:journals/ijon/LiYLM17}          \\
SD                   & 2012 & 0.483             & \cite{DBLP:journals/tmm/CongYL12}, \cite{DBLP:conf/cvip/MishraS16}          \\
VISON                & 2012 & 0.632             & \cite{DBLP:journals/prl/AlmeidaLT12}, \cite{MusselCirne2018}          \\
SFKD                 & 2013 & 0.531             & \cite{DBLP:journals/prl/MartinB13}, \cite{DBLP:journals/ijon/LiYLM17}           \\
KBKS                 & 2013 & 0.460             & \cite{DBLP:journals/tcsv/GuanWLDF13}, \cite{DBLP:conf/cvip/MishraS16}           \\
VSCAN                & 2013 & 0.713             & \cite{DBLP:conf/iciap/MahmoudIG13}, \cite{MusselCirne2018}          \\
HDPS                 & 2014 & 0.480             & \cite{rodriguez2014clustering}, \cite{Wu2017}          \\
OffMSR$_\text{a}$    & 2015 & 0.580             & \cite{DBLP:journals/pr/MeiGWWHF15}          \\
OffMSR$\text{m}$     & 2015 & 0.585             & \cite{DBLP:journals/pr/MeiGWWHF15}          \\
OffMSR               & 2015 & 0.569             & \cite{DBLP:journals/pr/MeiGWWHF15}          \\
VSQUAL               & 2016 & 0.636             & \cite{DBLP:conf/ciarp/CirneP14}, \cite{MusselCirne2018}          \\
VRHDPS               & 2017 & 0.630             & \cite{Wu2017}          \\
SUM-GAN$_\text{dpp}$ & 2017 & 0.728             & \cite{DBLP:conf/cvpr/MahasseniLT17}          \\
SUM-GAN$_\text{sup}$ & 2017 & 0.773             & \cite{DBLP:conf/cvpr/MahasseniLT17}          \\
SCVS                 & 2017 & 0.734             & \cite{DBLP:journals/ijon/LiYLM17}          \\
PFCVS                & 2017 & 0.609             & \cite{DBLP:conf/cvip/MishraS16}           \\
VISCOM               & 2017 & 0.721             & \cite{MusselCirne2018}          \\
AVS                  & 2018 & 0.648             & \cite{Kumar2018}         \\
EVS                  & 2018 & 0.631             & \cite{Kumar2018}         \\
ESVS                 & 2018 & 0.654             & \cite{Kumar2018}          \\
OPFSumm              & 2018 & 0.728             & \cite{martins2018opfsumm}          \\
MSKVS                & 2018 & 0.616             & \cite{DBLP:journals/jvcir/HannaneEA18}          \\
PICS                 & 2019 & 0.640             &  \cite{singh2019pics}         \\
SAEKMVS              & 2019 & 0.720             & \cite{DBLP:journals/jifs/MohanN19}          \\
KEGC                 & 2019 & 0.723             &  \cite{DBLP:journals/sivp/GharbiBZ19}         \\
CENTRIST VS          & 2019 & 0.600             & \cite{DBLP:journals/mta/BendraouES19}          \\ \hhline{====}
Proposed             & 2020 & \textbf{0.835}    &           \\ \hline
\end{tabular}%
}
\end{table}

For completeness we note that the f-measure values obtained via SalSum for each video in the VSUMM dataset as long as the selected number of keyframes are shown in the Table~\ref{tab:resultsAll}.

\begin{table}[H]
\centering
\caption{F-measure scores for each individual video of the VSUMM dataset obtained by the SalSum method. Mean score is in bold}
\label{tab:resultsAll}
\resizebox{0.75\textwidth}{!}{%
\begin{tabular}{cccccc}
\hline
Video & F-measure & Video & F-measure & Video                  & F-measure       \\ \hhline{======}
v21   & 0.6877    & v38   & 0.8301    & v55                    & 0.6367          \\
v22   & 0.9333    & v39   & 0.8686    & v56                    & 0.8878          \\
v23   & 0.7529    & v40   & 0.8950    & v57                    & 0.8572          \\
v24   & 0.8474    & v41   & 0.8552    & v58                    & 0.8679          \\
v25   & 0.8690    & v42   & 0.8984    & v59                    & 0.9018          \\
v26   & 0.8369    & v43   & 0.9102    & v60                    & 0.8756          \\
v27   & 0.7554    & v44   & 0.9778    & v61                    & 0.8214          \\
v28   & 0.9123    & v45   & 0.8659    & v62                    & 0.6500          \\
v29   & 0.9217    & v46   & 0.7917    & v63                    & 0.9179          \\
v30   & 0.7721    & v47   & 10.000    & v64                    & 0.8756          \\
v31   & 0.8637    & v48   & 0.9333    & v65                    & 0.8663          \\
v32   & 0.7036    & v49   & 0.7329    & v66                    & 0.7333          \\
v33   & 0.8742    & v50   & 0.8072    & v67                    & 0.7563          \\
v34   & 0.8742    & v51   & 0.8095    & v68                    & 0.7300          \\
v35   & 0.8722    & v52   & 0.8964    & v69                    & 0.7989          \\
v36   & 0.7000    & v53   & 0.8167    & v70                    & 0.8800          \\
v37   & 0.8000    & v54   & 0.8493    & \textit{\textbf{Mean}} & \textbf{0.8354} \\ \hline
\end{tabular}
}
\end{table}

\begin{figure*}[!h]
\centering
\includegraphics[width = 0.9\textwidth]{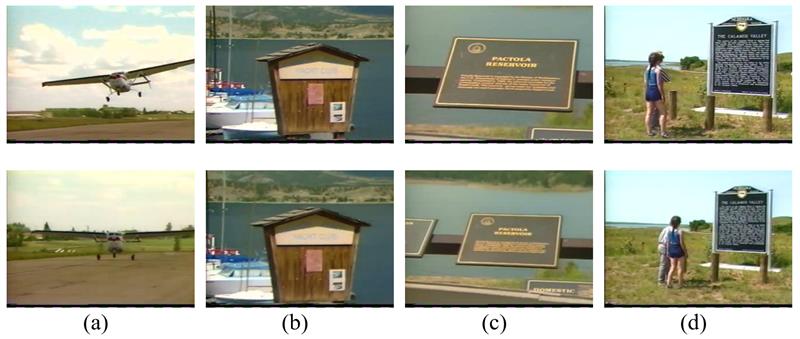}
\caption{Representative qualitative results. Keyframes selected by the user (first row) vs. keyframes selected by SalSum (second row) from video \#25 of the VSUMM dataset. (a) depicts airplane take-off, (b) a postal box and (c) a small sign, (d) a large sign.}
\label{fig: fig5}
\end{figure*}

A qualitative representation of the SalSUM performance is presented in Fig.~\ref{fig: fig5}. As it can be noticed, the selected keyframes by the proposed methodology are very closely related to and almost match the user-selected ones.

\section{Conclusions - Future Prospects}

In this work we introduce a novel VS method called SalSum. Salsum is based on both saliency information, deriving from human eye fixations estimates, in the form of saliency maps, provided by GAN model, and perceptual color cues. The incorporation of saliency and color information is achieved through the application of optical flow on saliency maps, capturing the spatial flow of attention in consecutive frames; while the perceptual color aspect is assessed by the quantization of  the hue channel, of HSV color space, on a frame. To determine the most effective way for the feature combination, a series of experiments were performed were SalSum performance was examined regarding both individual use of features and feature fusion using various techniques. The results obtained, indicate that a variance-based fusion between quantized hue of video frames and the optical flow of the corresponding saliency maps, provides the best video summarization performance. With this approach, the proposed method is able obtain perceptually relevant and intuitive video summaries, since any radical change both in the spatial attention flow and color information, is most likely to indicate a change in the video content. SalSum, when compared to several state-of-the-art VS approaches using the popular VSUMM dataset, outperformed them scoring an average f-measure score of 0.835 where the second best performing method achieved a score of 0.788.

As a future work, the incorporation of additional features will be considered, such as specified contexts within a video that interest individuals the most and/or human-based criteria like quality, quantity, beatifulness among others. Since human opinions may differ among different viewers, another interesting direction would be to study the GT selection in a video and compare humans themselves in order to achieve not the best VS that a user needs but the one that actually wants and suits his/her preferences.

\section*{Acknowledgments}
This research has been co-financed by the European Union and Greek national funds through the Operational Program Competitiveness, Entrepreneurship and Innovation, under the call RESEARCH - CREATE - INNOVATE (project code:T1EDK-02070)

{
\bibliographystyle{refbib}
\bibliography{SalSum_bib.bib}
}

\end{document}